\documentclass[sigconf, nonacm]{acmart}

\usepackage{amsmath}
\newcommand{\what}{\hat{w}}
\theoremstyle{acmplain}
\newtheorem{theorem}{Theorem}
\newtheorem{proposition}{Proposition}
\newtheorem{corollary}{Corollary}
\theoremstyle{acmdefinition}
\newtheorem{definition}{Definition}
\newtheorem{assumption}{Assumption}
\newcommand{\remarkhead}[2]{\par\textit{Remark #1 (#2).}\enspace}

\setcopyright{none}
\renewcommand\footnotetextcopyrightpermission[1]{}

\begin{document}

\title{FreshCast: Timely and Complementary Retrieval for Frozen Time Series Forecasters}

\author{Chao He}
\affiliation{\department{Aberdeen Institute of Data Science and Artificial Intelligence}\institution{South China Normal University}\city{Foshan}\country{China}}
\email{chaohe@m.scnu.edu.cn}

\author{Jianyu Xu}
\affiliation{\department{Aberdeen Institute of Data Science and Artificial Intelligence}\institution{South China Normal University}\city{Foshan}\country{China}}
\email{xujianyu@m.scnu.edu.cn}

\author{Xinyi Guo}
\affiliation{\department{Aberdeen Institute of Data Science and Artificial Intelligence}\institution{South China Normal University}\city{Foshan}\country{China}}
\email{20233801069@m.scnu.edu.cn}

\author{Ruiqi Liu}
\affiliation{\department{School of Artificial Intelligence}\institution{South China Normal University}\city{Foshan}\country{China}}
\email{liuruiqi@m.scnu.edu.cn}

\author{Haobin Ding}
\affiliation{\department{Aberdeen Institute of Data Science and Artificial Intelligence}\institution{South China Normal University}\city{Foshan}\country{China}}
\email{h.ding.262@gmail.com}

\author{Ruiqi He}
\affiliation{\department{Aberdeen Institute of Data Science and Artificial Intelligence}\institution{South China Normal University}\city{Foshan}\country{China}}
\email{202438012071@m.scnu.edu.cn}

\author{Yuntong Xie}
\affiliation{\department{Faculty of Humanities and Arts}\institution{Macau University of Science and Technology}\city{Macau}\country{China}}
\email{1250016321@student.must.edu.mo}

\author{Dongqing Song}
\authornote{Corresponding author.}
\affiliation{\department{School of Artificial Intelligence}\institution{South China Normal University}\city{Foshan}\country{China}}
\email{songdongqing@m.scnu.edu.cn}

\begin{abstract}
Retrieval-augmented forecasting for Web and online services supplements a forecaster with the continuations of historical segments that resemble the current context.
Most existing methods build the retrieval memory once from the training segment, leaving observations revealed after deployment unavailable as references, and generally do not calibrate how much retrieved information should influence a frozen forecaster.
We identify two key determinants of retrieval utility for a frozen forecaster: the timeliness of the retrieved history and its complementarity, i.e., whether the correction it induces aligns with the forecaster's residual errors; this alignment can shift between validation and deployment when the memory becomes stale.
We propose FreshCast, a plug-in retrieval framework for frozen forecasters: a non-parametric streaming memory keeps writing in new observations, relational kernel regression weights candidate readouts into a memory forecast, and its correction weight is calibrated in closed form on the validation segment.
Under a simplified generative model, we characterize the optimal combination gain through the second-order relation between forecaster error and memory correction, and show that periodic memory information can become redundant once the forecaster captures it.
Across seven datasets, ten forecasters, and three input lengths, FreshCast reduces average MSE for every forecaster and input length, by 14.0\% and 5.7\% at $L = 96$ and 720, and achieves lower average MSE than the evaluated retrieval-augmented and online updating baselines in their comparison settings.
Ablations show that post-training observations are a primary source of the gain: freezing the memory at the end of training removes most of it.
For a frozen forecaster, similarity alone is not enough: useful retrieval must stay timely and provide information that the forecaster still misses.
\end{abstract}


\maketitle

\section{Introduction}
\label{sec:intro}

Time-series forecasting supports traffic scheduling, electricity load management, weather forecasting, and capacity planning in Web and online services.
Deep forecasters keep improving on standard benchmarks~\cite{zhou2021informer, wu2021autoformer, zeng2023dlinear, nie2023patchtst, wu2023timesnet, liu2024itransformer}, but they typically predict from a fixed-length look-back window, so history outside the window acts only indirectly, through parameters learned in training.
Retrieval-augmented forecasting provides a complementary mechanism by retrieving historical segments that resemble the current context and using their continuations as forecasting references, as in RAFT~\cite{han2025raft}, PFRP~\cite{du2026pfrp}, and GTR~\cite{cao2026gtr}.
Like retrieval-augmented AI for Web search, recommendation, and generation, it draws on an external store of past data, valued in retrieval-augmented generation because it can be updated without retraining~\cite{lewis2020rag}; a key distinction for time series is that the available history continues to grow after deployment, while the deployed forecaster may remain fixed.
Existing work on this paradigm has largely focused on how history is represented, how segments are matched, and how retrieved information is fused, with matching quality often taken as a proxy for usefulness.

\begin{figure*}[t]
  \centering
  \includegraphics[width=\textwidth]{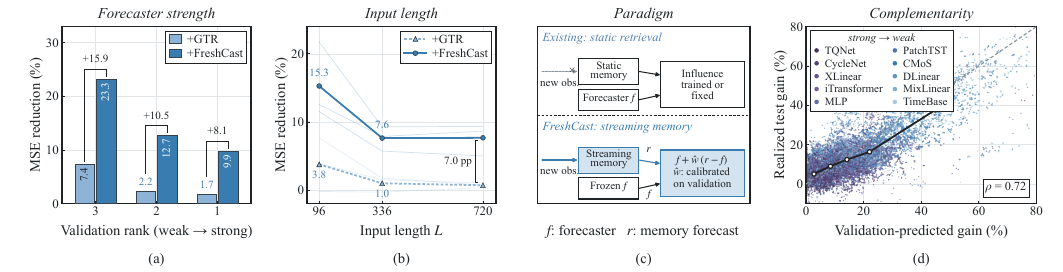}
  \caption{Retrieval utility for frozen forecasters. (a) The gain of the same retrieval scheme (GTR) shrinks on more accurate forecasters (MLP, DLinear, and iTransformer ranked by validation error; brackets: gap in points). (b) It also shrinks as the input length $L$ grows. (c) Existing retrieval keeps a static memory with an influence trained or fixed; FreshCast streams the memory and calibrates the weight on validation data. (d) The validation-estimated optimal combination gain tracks the realized test gain ($L = 96$, ten forecasters; Spearman 0.72).}
  \label{fig:intro}
\end{figure*}

Our starting point is a gap between matching quality and forecasting gain: a retrieved segment can resemble the current context in shape while its continuation no longer matches the current state, and the gain from even a valid reference also depends on what the forecaster already captures.
We focus on two common design choices that contribute to this gap (Figure~\ref{fig:intro}(c), Section~\ref{sec:related}).
First, the retrieval memory is typically fixed when training ends, so observations revealed after deployment do not enter the candidate set.
Second, the degree to which retrieved information influences the forecast is typically learned on the training data or fixed by design, and is generally not calibrated for a frozen forecaster; because the forecaster has been fitted on the training segment, estimating retrieval utility from the same data can underestimate the value of a correction (Section~\ref{sec:ablation}).
Observation~2 below concerns the first choice, and Observations~1 and~3 the second.

\paragraph{Observation 1: retrieval gains shrink as periodic context enters the look-back window.}
The same retrieval scheme yields smaller gains on forecasters with lower validation error (Figure~\ref{fig:intro}(a)).
With the forecaster fixed, extending the look-back window from 96 to 720 steps reduces the average error reduction of GTR on MLP, DLinear, and iTransformer, the forecasters evaluated at all three lengths, from 3.78\% to 0.71\% (Figure~\ref{fig:intro}(b)), and on Electricity, whose main period is one week, from 11.84\% to 1.71\%.
Since GTR indexes its references by periodic position, these results suggest that a substantial part of its gain comes from periodic information that is unavailable to a short look-back window; once that structure becomes accessible within the input, this source of retrieval gain diminishes.

\begin{figure*}[t]
  \centering
  \includegraphics[width=\textwidth]{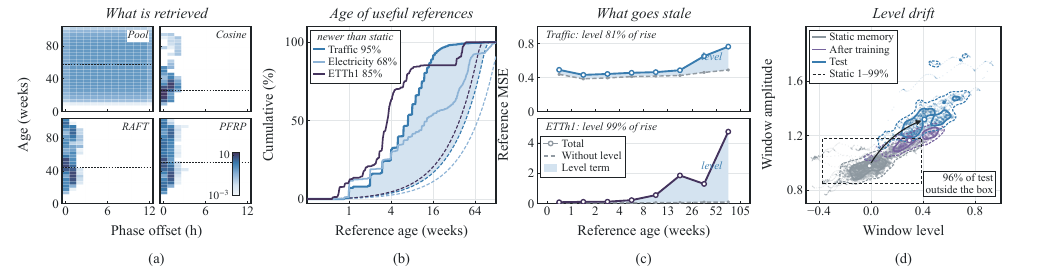}
  \caption{Similar shapes can carry stale levels. (a) Traffic: phase offset and age of candidates and of segments retrieved by cosine similarity, RAFT, and PFRP. (b) Age of the ten most useful references in hindsight (solid) and of candidates (dashed); legend: share newer than the static memory. (c) Error of same-phase value readouts by age (solid), without the level term (dashed), and the level term (shaded); top: Traffic, bottom: ETTh1. (d) Level and amplitude of Traffic windows in the static memory, after training, and at test queries (contours: 50\%, 90\%; box: 1--99\% range of the static memory).}
  \label{fig:retrieval}
\end{figure*}

\paragraph{Observation 2: references with similar shapes can carry outdated levels.}
On Traffic, 95.8\%--99.8\% of the segments retrieved by five rules, including those of RAFT and PFRP, lie near the query's time of day, against 12.5\% of the candidate set (Figure~\ref{fig:retrieval}(a)), yet in a post hoc evaluation 95\% of the most useful segments are newer than the newest segment of a static memory built from the training segment (Figure~\ref{fig:retrieval}(b)).
Error growth with age comes mainly from the level term of the reference error (Figure~\ref{fig:retrieval}(c)), and test windows drift outside the level and amplitude range of training windows (Figure~\ref{fig:retrieval}(d)).
These results suggest that historical shape can remain informative after its absolute level or amplitude becomes stale: across datasets, the win rate of a seasonal analog's shape readout over its value readout rises with age and passes one half at about 2--4 weeks, except on Traffic, where level gaps stay small.
When the level itself varies seasonally, older segments from the corresponding season can become useful again (Appendix~\ref{app:limitations}).

\paragraph{Observation 3: complementarity on validation data correlates with the realized test gain.}
We characterize complementarity through the relation between the forecaster error and the memory-induced correction: the correction $r - f$ is useful when it aligns with the residual error $y - f$.
Over all evaluated forecasters and datasets, the validation-estimated optimal combination gain correlates with the realized test gain (Spearman 0.72, 0.52, and 0.47 at $L = 96$, 336, and 720; Figure~\ref{fig:intro}(d)), complementarity itself correlates best among validation indicators (Appendix~\ref{app:theory-checks}), and with the same memory, more accurate forecasters tend to exhibit lower complementarity and smaller gains, consistent with Observation~1.

These observations prompt the following question: for a forecaster that is trained and then kept fixed, how much of the retrieval gain comes from information absent at training time, and how much from information the forecaster has not captured?
The former concerns timeliness, whether the retrieved history still reflects the current state, and calls for a memory that keeps updating during deployment (Observation~2); the latter concerns the complementarity between the memory correction and the forecaster, and calls for adopting the correction according to its relation to the forecaster's residuals (Observations~1 and~3).
These two key determinants of retrieval utility for frozen forecasters interact: if the memory becomes stale after calibration, the usefulness of the correction can shift between validation and deployment (Theorem~\ref{thm:safety}, Section~\ref{sec:ablation}).

To this end, we propose FreshCast, a plug-in framework built around these two determinants (Figure~\ref{fig:arch}).
It casts retrieval as a residual correction $\hat{y} = f + \what\,(r - f)$ on a frozen forecaster, where the memory forecast $r$ is built without forecaster outputs and can thus serve a wide range of trained forecasters without modifying their parameters.
For timeliness, a \textbf{streaming memory} writes in post-deployment observations at every forecast origin; seasonal analogs from the same phase in recent periods are more likely to retain useful absolute values and are read by \textbf{value}, whereas shape analogs from other times may sit at another level and are read by \textbf{shape} after re-anchoring to the query's level and amplitude.
\textbf{Relational kernel regression} weights candidate readouts by their relation to the query (shape similarity, age, phase, level gap, and scale mismatch) and preserves complementarity at deployment better than an untrained rule selected per channel (Section~\ref{sec:kernel}).
For complementarity, \textbf{combination-weight calibration} estimates $\what$ on the validation segment: under squared error, the optimal linear coefficient for the residual correction has a closed form determined by the alignment and magnitude of the correction relative to the forecaster error (Proposition~\ref{prop:gain}).
The calibration itself introduces no trainable parameters, and a zero estimate recovers the original forecaster.
Keeping the memory updated reduces one important source of mismatch between validation-time calibration and deployment-time retrieval (Theorem~\ref{thm:safety}).

Our main contributions are as follows.
\begin{itemize}
  \item \textbf{Analysis.} We identify two key determinants of retrieval utility for frozen forecasters: the timeliness of retrieved history and the alignment between memory-induced corrections and the forecaster's residuals, which we call complementarity. We further show that memory staleness can shift this relationship between validation and deployment.
  \item \textbf{Method.} We propose FreshCast, a plug-in framework that keeps the forecaster frozen, refreshes a streaming memory, forms a memory forecast by relational kernel regression, and calibrates the residual correction in closed form.
  \item \textbf{Theory and experiments.} Under a simplified model with level and amplitude variation and periodic structure, we characterize the achievable linear-combination gain, analyze how memory age and look-back length affect retrieval utility, and bound the probability that validation-calibrated combination underperforms the original forecaster. \emergencystretch=1em Across ten forecasters, seven datasets, and three input lengths, FreshCast consistently reduces average forecasting error and achieves lower MSE than the evaluated re\-trie\-val-aug\-ment\-ed baselines in each baseline's comparison setting, as well as six online updating methods on the same forecaster. Freezing the memory at the end of training removes most of the improvement, showing that post-training observations are a primary source of the gain, and giving RAFT the same observations does not close the gap.
\end{itemize}

\section{Related Work}
\label{sec:related}

\paragraph{Deep time-series forecasting.}
Architectures for long-term forecasting range from Transformers~\cite{nie2023patchtst, liu2024itransformer} to linear and MLP models~\cite{zeng2023dlinear, huang2025timebase, si2025cmos, ma2026mixlinear, chen2026xlinear} and models that explicitly model periodicity~\cite{lin2024cyclenet, lin2025tqnet}; despite their different architectures, these forecasters typically predict from a fixed-length look-back window.
For such a frozen forecaster, new observations inside the window still enter the forecast, but observations outside it act only through information encoded during training, and post-training history remains inaccessible; we supply it from outside, leaving the forecaster unchanged.

\paragraph{Retrieval-augmented time-series forecasting.}
Retrieval-augmented methods retrieve historical segments similar to the current context and use their continuations as forecasting references~\cite{han2025raft}.
Among methods that make retrieval part of the model, existing work improves how history is represented (PFRP, with a contrastively trained encoder of training windows~\cite{du2026pfrp}), how references are organized (GTR, with a learned table indexed by periodic position~\cite{cao2026gtr}), how they are selected (SARAF, which notes that highly similar past segments need not have similar futures and diversifies them by stationarity~\cite{zhou2026saraf}), and how they are fused (RAFT, with a fusion module trained jointly with the forecaster~\cite{han2025raft}).
These improvements raise accuracy on standard benchmarks; in these methods the retrieval memory is constructed from the training segment, so observations revealed after deployment do not enter the candidate set.
Other methods attach retrieval outside an unchanged forecaster.
RAF and TS-RAG feed retrieved series to foundation models without fine-tuning~\cite{tire2026raf, ning2025tsrag}; RAF requires the retrieval memory to contain only time steps strictly earlier than the forecast window of the query, but constructs this memory once when the data are split~\cite{tire2026raf}.
kNN-MTS sets the mixing weight by retrieval distance, with a datastore built offline on the training set~\cite{zhang2025knnmts}, and CRAFT uses a small fixed coefficient~\cite{kang2026craft}; in neither case is the mixing coefficient calibrated from the alignment between the retrieval-induced correction and the residual errors of the attached forecaster, that is, from complementarity.
Weighting two forecasts and testing forecast encompassing are classical tools of forecast combination, and StreamWNN updates a nearest-neighbor forecaster incrementally on streaming data~\cite{melgar2023streamwnn}.
FreshCast enforces the causal constraint of RAF at every forecast origin and lets the memory grow with new observations to stay timely, and it calibrates the combination weight by how much the memory correction explains the forecaster's residuals.

\paragraph{Distribution shift and online forecasting.}
Non-stationary forecasting methods mitigate distribution variation through normalization or learned transformations of input statistics~\cite{kim2022revin, liu2022nstransformer, fan2023dishts, liu2023san, ye2024fan}.
Test-time adaptation and online forecasting methods incorporate ground truth arriving during deployment by updating the forecaster, an adaptation module, or online-fitted predictive components~\cite{kim2025tafas, pham2023fsnet, zhang2023onenet, chen2024solid, lau2025dsof, zhao2025proceed, chen2025sttc, lee2025elf}.
Some keep the forecaster frozen: PETSA and COSA update small calibration modules at test time~\cite{medeiros2025petsa, im2026cosa}, $\delta$-Adapter learns small, bounded input and output corrections that can be updated online, its output correction acting where ours does~\cite{liang2026forecast}, and DynaME fits experts in closed form on period-lagged samples at every step~\cite{hong2026dyname}.
FreshCast updates no parameter during deployment: new observations enter a non-parametric memory as raw values, and the only quantity fitted to the forecaster, one combination weight per dataset and horizon, is estimated in closed form before deployment.

\section{Method}
\label{sec:method}

\begin{figure*}[t]
  \centering
  \includegraphics[width=0.85\textwidth]{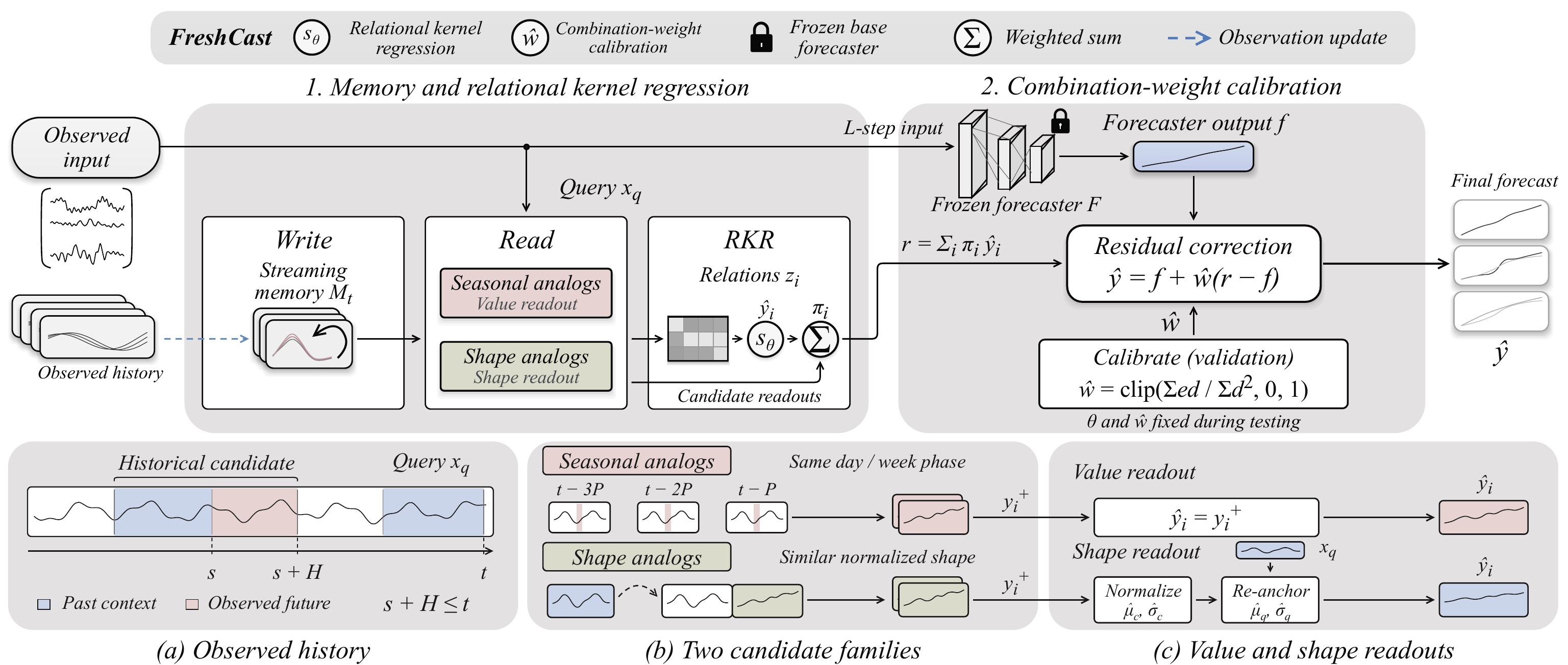}
  \caption{Overview of FreshCast. Write: the streaming memory stores post-deployment observations while the forecaster stays frozen. Read: seasonal analogs by value, shape analogs by shape. RKR: candidates are weighted by their relation to the query into $r$. Calibrate: $\what$ is estimated in closed form on the validation segment. Bottom: (a) causally admissible windows, (b) candidate families, (c) value and shape readouts (schematic).}
  \label{fig:arch}
\end{figure*}

\subsection{Problem Setting}
\label{sec:setting}

A series has $C$ channels, each standardized with its training-segment statistics so that retrieval distances are comparable across channels.
At forecast origin $t$, a frozen forecaster $F$ maps a look-back window of length $L$ to the next $H$ steps.
For one channel, we write $y$ for the series, $f \in \mathbb{R}^H$ for the forecaster output, and $x_q$ for the query (the latest $L_q$ steps), with mean $\hat{\mu}_q$ and standard deviation $\hat{\sigma}_q$.
A candidate is a past window ending at time $s$ with its next $m$ steps.

\begin{definition}[Causal admissibility]
\label{def:causal}
A candidate is causally admissible at origin $t$ if and only if $s + m \le t$.
\end{definition}

\begin{definition}[Streaming and static memory]
\label{def:memory}
A streaming memory writes in newly revealed observations at every origin; its candidate set $\mathcal{M}_t$ holds the causally admissible candidates up to $t$ (part of them under bounded storage; Appendix~\ref{app:impl}). A static memory is a streaming memory whose writing stops at the training cutoff.
\end{definition}

We aim to construct, without modifying $F$ and from observations up to $t$ only, a memory forecast $r \in \mathbb{R}^H$ and a combination weight $w$ that give
\begin{equation}
  \hat{y} = f + w\,(r - f)
  \label{eq:output}
\end{equation}
a small squared error.
The memory forecast is built per channel; $w$ is a generic weight, and FreshCast uses the estimate $\what$ of Section~\ref{sec:calibration}, shared across the channels of a dataset for each forecaster and horizon.
Ground-truth futures are used only after they arrive: as training labels and loss scale of relational kernel regression, for calibration and model selection on the validation segment, and for test evaluation.

\subsection{Structure Overview}
\label{sec:overview}

Figure~\ref{fig:arch} shows the architecture of FreshCast, which keeps the forecaster frozen and moves all deployment-time updates into a non-parametric memory: (i) the \emph{streaming memory} (Section~\ref{sec:memory}) writes in observations revealed after deployment, keeping the retrieved history timely, and supplies two families of references, seasonal and shape analogs; (ii) \emph{relational kernel regression} (Section~\ref{sec:kernel}) combines candidate readouts into $r$ by their relation to the query; and (iii) \emph{combination-weight calibration} (Section~\ref{sec:calibration}) sets how much of the correction $r - f$ is adopted by how much it explains the forecaster's residuals.
In summary,
\begin{equation}
  \begin{aligned}
    \mathcal{M}_t &= \mathrm{Write}(\mathcal{M}_{t-1}, y_t), & \mathcal{C}_t &= \mathrm{Read}(\mathcal{M}_t, x_q), \\
    r &= \mathrm{RKR}_\theta(\mathcal{C}_t, x_q), & \hat{y} &= f + \what\,(r - f),
  \end{aligned}
  \label{eq:pipeline}
\end{equation}
where $\mathcal{C}_t$ denotes the candidates after reading; during deployment only $\mathcal{M}_t$ changes, while $\theta$ and $\what$ are fixed beforehand.

\subsection{Streaming Memory and Two Families of References}
\label{sec:memory}

To keep the retrieved history timely, the memory of FreshCast keeps writing in observations (Observation~2), so the newest reference always has age $H$ (Corollary~\ref{cor:static}; stored contents in Appendix~\ref{app:impl}).
Following classical analog forecasting, the two families of references are read according to their source.
A continuation $y_i^{+}$ is read either by value, $\hat{y}_i = y_i^{+}$, or by shape, standardized with the mean $\hat{\mu}_c$ and standard deviation $\hat{\sigma}_c$ of the candidate's past window and re-anchored to the query:
\begin{equation}
  \hat{y}_i = \hat{\mu}_q + \frac{\hat{\sigma}_q}{\hat{\sigma}_c}\,\bigl(y_i^{+} - \hat{\mu}_c\bigr),
  \label{eq:reanchor}
\end{equation}
where $\hat{y}_i \in \mathbb{R}^H$ is the readout of candidate $i$.
\emph{Seasonal analogs} take the same phase $1$ to $J$ daily and weekly periods back within the most recent $K$ days.
In a streaming memory they are as young as one period and thus more likely to retain the current level (Proposition~\ref{prop:staleness}, Corollary~\ref{cor:static}), so they are read by value, which gives 0.1\%--0.2\% lower error than offering both readouts (Appendix~\ref{app:theory-checks}).
Once periodic structure enters the look-back window, what remains to be added under the condition of Proposition~\ref{prop:chain} is mainly state information, which can come from segments at other times with similar trajectories: the \emph{shape analogs}, the $N$ windows whose shape vectors are most cosine-similar to the query, read by Eq.~\eqref{eq:reanchor}.
When the local level and amplitude are approximately constant, their error has no first-order level term (Proposition~\ref{prop:staleness}).
Timeliness thus simplifies reading: only shape analogs, which differ from the query in level, need re-anchoring, and they help more where events recur at irregular times (Appendix~\ref{app:synthetic}).

\subsection{Relational Kernel Regression}
\label{sec:kernel}

Observation~2 suggests that shape similarity alone cannot tell which of the roughly two hundred candidates per query can be trusted, as trust relates to how a candidate stands relative to the current state.
An untrained rule selected per channel matches a learned weighting in validation complementarity, but its complementarity drops more at deployment and the calibrated weight becomes misaligned (Theorem~\ref{thm:safety}(iii), Appendix~\ref{app:theory-checks}); we therefore learn the weighting.
Relational kernel regression takes the form of classical kernel regression, with a kernel learned from relation descriptors instead of a fixed bandwidth.
The descriptor $z_i$ covers the candidate's family and two phase distances (periodic position), shape similarity (match quality), age, level gap, and scale ratio (consistency with the current state), plus training-segment series descriptors (Appendix~\ref{app:impl}).
Since the gain is largest when shape analogs and nonlinear scoring are combined (Section~\ref{sec:ablation}), the weights and the memory forecast are
\begin{equation}
  \pi_i = \frac{\exp\bigl(\beta\, s_\theta(z_i)\bigr)}{\sum_{j} \exp\bigl(\beta\, s_\theta(z_j)\bigr)}, \qquad r = \sum_i \pi_i\, \hat{y}_i,
  \label{eq:kernel}
\end{equation}
where $z_i \in \mathbb{R}^{11}$, $s_\theta$ is a linear term plus an $11 \to 64 \to 64 \to 1$ multilayer perceptron, and $\beta$ is a global inverse temperature.
Relational kernel regression only assigns weights, so $r$ lies in the convex hull of the candidate readouts.
The loss is normalized by the validation error $\varepsilon_c$ of the untrained rule on each channel and clipped, so that channels with different error magnitudes are comparable:
\begin{equation}
  \ell = \min\!\left( \frac{\lVert r - y_{t+1:t+H} \rVert_2^2}{H\, \varepsilon_c},\; c_{\max} \right).
  \label{eq:loss}
\end{equation}
One model per horizon is trained on training-segment samples and shared across the sampled channels of all datasets.

\subsection{Combination-Weight Calibration}
\label{sec:calibration}

An accurate memory forecast need not complement the forecaster, so we calibrate the combination weight to adopt the correction as far as it explains the forecaster's residuals (Observations~1 and~3).
With $e = y - f$ and $d = r - f$, the mean squared error of $e - w\,d$ is minimized at $w^{*} = \mathbb{E}[e\,d] / \mathbb{E}[d^{2}]$ (Proposition~\ref{prop:gain}), which depends on both the alignment and the magnitude of the correction, as in the classical optimal combination of two forecasts.
We estimate it on the validation segment, which the forecaster has not fitted, and clip it to $[0, 1]$, since a training-segment estimate can underestimate the value of the correction (Section~\ref{sec:ablation}):
\begin{equation}
  \what = \operatorname{clip}\!\left( \frac{\sum_{i} e_i\, d_i}{\sum_{i} d_i^{2}},\; 0,\; 1 \right)
  \label{eq:what}
\end{equation}
for each forecaster, dataset, and horizon, with sums over validation origins (100 per channel; Appendix~\ref{app:impl}), channels, and forecast steps to enlarge the estimation sample (Theorem~\ref{thm:safety}); the test segment reuses $\what$.
Before clipping, $\what$ is the coefficient of regressing $e$ on $d$ in the classical forecast-encompassing test, and a non-positive value means that the forecaster encompasses the memory forecast on the validation segment.
The weight $\what$ thus interpolates between the frozen forecaster ($\what = 0$) and the memory forecast ($\what = 1$), falling back to the original forecaster at $\what = 0$, and the calibration introduces no trainable parameters.

\subsection{Training and Inference}
\label{sec:training}

Within FreshCast, only relational kernel regression (about $5.0 \times 10^{3}$ parameters) is trained, by Eq.~\eqref{eq:loss} alone and without forecaster outputs, so one model serves all forecasters and attaching a new forecaster needs only Eq.~\eqref{eq:what}.
The number of candidates per query is fixed and does not grow with deployment time (costs in Section~\ref{sec:efficiency}).

\section{Theoretical Analysis}
\label{sec:theory}

This section explains the three observations under a simplified, additive, per-channel generative model and delimits the conditions under which each component applies (proofs in Appendix~\ref{app:proofs}, checks in Appendix~\ref{app:theory-checks}).

\paragraph{Setting.}
Fix one channel and let $t$ be the forecast origin.
\begin{assumption}
\label{asm:structure}
The series satisfies $y_t = m_t + a_t\, \psi(\varphi_t) + \varepsilon_t$, where $\psi$ has period $P$ with zero mean and unit mean square over one period, $\varepsilon_t$ is i.i.d.\ with variance $\sigma_\varepsilon^2$ and independent of $(m, a)$, and the increments of level and amplitude are independent of the phase.
\end{assumption}
Let $D(\tau) = \mathbb{E}[(m_t - m_{t-\tau})^2]$ and $A(\tau) = \mathbb{E}[(a_t - a_{t-\tau})^2]$ be the level and amplitude mismatches at age $\tau$, $f = F(X_L)$, and $r = R(\mathcal{M}_t, X_L)$ for any mapping $R$, relational kernel regression being one instance.
With $e = y - f$ and $d = r - f$, let $S_{ee}$, $S_{ed}$, and $S_{dd}$ be their second moments, and define the uncentered correlation $\rho = S_{ed} / \sqrt{S_{ee} S_{dd}}$, the complementarity $\kappa = S_{ed} / S_{ee}$, and the divergence $\delta = S_{dd} / S_{ee}$; $\kappa > 0$ means that the correction aligns with the residual (Section~\ref{sec:intro}).

\begin{proposition}[Redundancy of periodic information; Observation~1]
\label{prop:chain}
Split the memory into a periodic part $M_p$ and a state part $M_s$ after the end of training. Then $I(Y; \mathcal{M} \mid X_L) = I(Y; M_p \mid X_L) + I(Y; M_s \mid X_L, M_p)$, and if $Y$ and $M_p$ are conditionally independent given $X_L$, the first term vanishes.
\end{proposition}

\remarkhead{1}{Periodic context}
Conditional independence requires the forecaster to have absorbed the periodic information in its window, which $L \ge P$ alone does not guarantee.
The proposition only says that the periodic part of the gain bound can vanish, in line with the periodic-tiling control (Appendix~\ref{app:theory-checks}); whether the state part, which shape analogs supply for long inputs (Section~\ref{sec:memory}), is positive is an empirical question (Sections~\ref{sec:generality} and~\ref{sec:causal}).

\begin{proposition}[Timeliness; Observation~2]
\label{prop:staleness}
Let a candidate share the phase of the query and have age $\tau$.
(i) The error of the value readout $y^{+}$ is $\mathbb{E}[(y - y^{+})^2] = D(\tau) + A(\tau) + 2\sigma_\varepsilon^2$.
(ii) If level and amplitude are approximately constant over one look-back window plus one horizon, the error of the shape readout contains, to first order, no level term, and amplitude differences mainly rescale the noise.
\end{proposition}

\begin{corollary}[Memory age]
\label{cor:static}
With training ending at $T_{\mathrm{tr}}$, the minimum age of segments in a static memory, $t - T_{\mathrm{tr}} + H$, grows linearly with deployment time, whereas in a streaming memory it is always $H$. If the level process is a stationary, mixing Markov process, the information about the current level carried by value readouts decays accordingly.
\end{corollary}

\remarkhead{2}{Readouts}
The premise of (ii) is local and does not involve the level difference between segment and query.
The level term accounts for 81\%--99\% of the age-related error growth of value-read seasonal analogs (Appendix~\ref{app:theory-checks}), consistent with (i); this does not imply that shape is preserved, and when the level oscillates seasonally, segments from a year earlier can become useful again.

\begin{proposition}[Combination gain; Observation~3]
\label{prop:gain}
For any weight $w$, $\mathrm{MSE}(w) = S_{ee}(1 - \rho^2) + S_{dd}(w - w^{*})^2$ with $w^{*} = S_{ed} / S_{dd}$. If $0 \le w^{*} \le 1$, the largest error reduction relative to the forecaster is $g^{*} = \rho^2 = \kappa^2 / \delta$; if $S_{ed} \le 0$, the optimal weight is $0$.
\end{proposition}

\remarkhead{3}{Retrieval as an explanation of residuals}
Proposition~\ref{prop:gain} is the classical optimal combination of two forecasts; we use it to express retrieval utility as a quantity estimable before deployment, namely how much the correction explains the residuals of the frozen forecaster.
The achievable gain is characterized by the second-order relation between forecaster error and memory correction, depends on both $\kappa$ and $\delta$, and yields the closed form and clipping of Eq.~\eqref{eq:what}; its validation estimate correlates with the test gain (Appendix~\ref{app:theory-checks}).

\begin{theorem}[Harm of the validation-calibrated weight]
\label{thm:safety}
Let $0 < w^{*} \le 1$ and let $\what$ be estimated from $n$ validation samples by Eq.~\eqref{eq:what}.
(i) $\mathrm{MSE}(\what) - S_{ee} = S_{dd}[(\what - w^{*})^2 - w^{*2}]$, so $\what$ is worse than the forecaster if and only if $|\what - w^{*}| > w^{*}$.
(ii) If the validation samples are i.i.d.\ and follow the test distribution, the probability of this event is bounded by a term that decays exponentially in $n$, whose leading part decreases as $n\kappa^2$ grows.
(iii) Part (i) holds with $S_{dd}$ and $w^{*}$ replaced by their test counterparts: the harm at deployment is determined jointly by the estimation error and the shift between the validation and test optima.
\end{theorem}

\remarkhead{4}{Interaction of timeliness and complementarity}
Part (iii) characterizes how the two key determinants interact: a stale memory can move the test optimum away from the validation estimate. With a static memory, the validation $\what$ is too high in 17--26 of 28 settings (Appendix~\ref{app:theory-checks}), as in the ablation cell where both axes are unfavorable (Section~\ref{sec:ablation}), which is also why $\what$ is estimated on the validation segment and pooled across channels, unlike the per-channel model above.
The i.i.d.\ premise of (ii) does not strictly hold for time series (observed harm rates in Appendix~\ref{app:theory-checks}), and falling back to the forecaster at $\what = 0$ is not a per-sample reliability guarantee (Appendix~\ref{app:limitations}).

\setcounter{dbltopnumber}{2}
\section{Experiments}
\label{sec:experiments}

\begin{table*}[t]
\caption{Test MSE of the ten forecasters alone (Base) and with FreshCast at $L = 96$ (mean over four horizons); lower block: forecasters proposed in 2025--2026.}
\label{tab:generality}
\footnotesize
\setlength{\tabcolsep}{0pt}
\begin{tabular*}{\textwidth}{@{\extracolsep{\fill}}l*{10}{c}}
\toprule
 & \multicolumn{2}{c}{MLP} & \multicolumn{2}{c}{DLinear} & \multicolumn{2}{c}{PatchTST} & \multicolumn{2}{c}{iTransformer} & \multicolumn{2}{c}{CycleNet} \\
\cmidrule(lr){2-3}\cmidrule(lr){4-5}\cmidrule(lr){6-7}\cmidrule(lr){8-9}\cmidrule(lr){10-11}
Dataset & Base & +FreshCast & Base & +FreshCast & Base & +FreshCast & Base & +FreshCast & Base & +FreshCast \\
\midrule
ETTh1 & 0.449 & 0.404 & 0.462 & 0.443 & 0.461 & 0.404 & 0.472 & 0.406 & 0.450 & 0.404 \\
ETTh2 & 0.374 & 0.368 & 0.560 & 0.373 & 0.381 & 0.375 & 0.387 & 0.376 & 0.378 & 0.373 \\
ETTm1 & 0.383 & 0.328 & 0.403 & 0.339 & 0.383 & 0.329 & 0.413 & 0.335 & 0.380 & 0.328 \\
ETTm2 & 0.279 & 0.250 & 0.347 & 0.251 & 0.285 & 0.251 & 0.290 & 0.252 & 0.270 & 0.248 \\
Weather & 0.258 & 0.222 & 0.267 & 0.226 & 0.262 & 0.223 & 0.270 & 0.225 & 0.246 & 0.220 \\
Electricity & 0.190 & 0.158 & 0.210 & 0.159 & 0.200 & 0.159 & 0.175 & 0.153 & 0.168 & 0.155 \\
Traffic & 0.459 & 0.393 & 0.626 & 0.400 & 0.492 & 0.397 & 0.424 & 0.384 & 0.484 & 0.393 \\
\midrule
 & \multicolumn{2}{c}{TQNet} & \multicolumn{2}{c}{XLinear} & \multicolumn{2}{c}{MixLinear} & \multicolumn{2}{c}{TimeBase} & \multicolumn{2}{c}{CMoS} \\
\midrule
ETTh1 & 0.444 & 0.404 & 0.433 & 0.402 & 0.447 & 0.400 & 0.463 & 0.402 & 0.475 & 0.399 \\
ETTh2 & 0.378 & 0.373 & 0.372 & 0.370 & 0.384 & 0.379 & 0.405 & 0.386 & 0.394 & 0.376 \\
ETTm1 & 0.378 & 0.329 & 0.374 & 0.329 & 0.422 & 0.333 & 0.427 & 0.334 & 0.401 & 0.332 \\
ETTm2 & 0.278 & 0.249 & 0.272 & 0.249 & 0.288 & 0.254 & 0.291 & 0.254 & 0.286 & 0.253 \\
Weather & 0.242 & 0.220 & 0.238 & 0.217 & 0.275 & 0.226 & 0.253 & 0.222 & 0.253 & 0.223 \\
Electricity & 0.165 & 0.150 & 0.172 & 0.154 & 0.224 & 0.160 & 0.227 & 0.160 & 0.203 & 0.160 \\
Traffic & 0.445 & 0.389 & 0.465 & 0.394 & 0.637 & 0.400 & 0.681 & 0.400 & 0.552 & 0.399 \\
\bottomrule
\end{tabular*}
\end{table*}

\begin{figure*}[t]
  \centering
  \includegraphics[width=\textwidth]{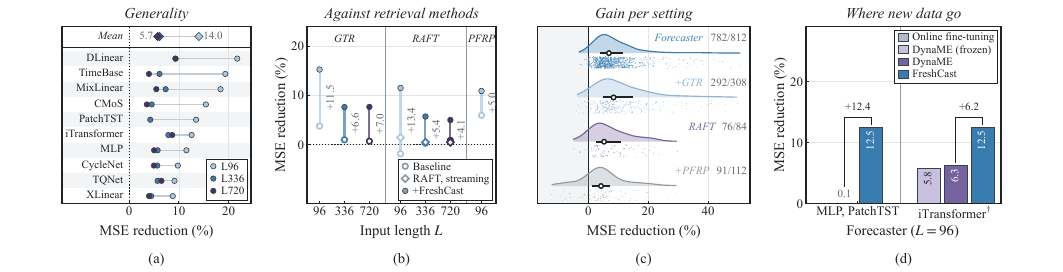}
  \caption{FreshCast lowers the average error of every evaluated forecaster. (a) MSE reduction of FreshCast per forecaster and input length (top row: mean). (b) MSE reduction of retrieval-augmented baselines (hollow; GTR without PatchTST; diamonds: streaming RAFT) and FreshCast (filled); numbers: gap in points. (c) Per-setting MSE reduction of FreshCast relative to the forecaster, +GTR, RAFT, and +PFRP (medians and interquartile ranges), with the number of positive settings. (d) MSE reduction at $L = 96$ when new observations go into parameters (online fine-tuning, DynaME) or into the memory (FreshCast).}
  \label{fig:results}
\end{figure*}

\subsection{Experimental Setup}
\label{sec:exp-setup}

\paragraph{Data and forecasters.}
We use seven long-term forecasting benchmarks from the electricity, traffic, and weather domains (ETTh1, ETTh2, ETTm1, ETTm2, Weather, Electricity, and Traffic~\cite{zhou2021informer, wu2021autoformer}), split chronologically into training, validation, and test segments (12/4/4 months for ETT, 7:1:2 otherwise; statistics in Appendix~\ref{app:impl}).
The ten forecasters span MLP, linear, Transformer, and periodicity-modeling architectures: MLP, DLinear~\cite{zeng2023dlinear}, PatchTST~\cite{nie2023patchtst}, iTransformer~\cite{liu2024itransformer}, and CycleNet~\cite{lin2024cyclenet}, and the 2025--2026 models TQNet~\cite{lin2025tqnet}, XLinear~\cite{chen2026xlinear}, MixLinear~\cite{ma2026mixlinear}, TimeBase~\cite{huang2025timebase}, and CMoS~\cite{si2025cmos}.
Each is trained with its official configuration and frozen, and FreshCast is attached to its checkpoint without modifying any parameter, so that changes in error can be attributed to the memory branch and its combination.
We use the input lengths of the official configurations, $L \in \{96, 336, 720\}$ (PatchTST only at 96 and 336; Appendix~\ref{app:impl}), and $H \in \{96, 192, 336, 720\}$.
Two pretrained foundation models, Chronos-Bolt~\cite{ansari2024chronos} and Sundial~\cite{liu2025sundial}, serve as zero-shot frozen forecasters with a 512-step context.

\paragraph{Baselines.}
We compare FreshCast with GTR~\cite{cao2026gtr}, RAFT~\cite{han2025raft}, and PFRP~\cite{du2026pfrp} under the supported comparison protocols below.
GTR provides implementations for MLP, DLinear, PatchTST, and iTransformer; RAFT, a complete model with a shallow linear backbone, is compared with the structurally closest MLP+FreshCast; and PFRP, which has only a univariate implementation, is compared on the four forecasters and target channels of its paper at $L = 96$.
Baselines otherwise use the settings of their papers or official scripts. Six online updating methods, DynaME~\cite{hong2026dyname} (tuned on validation data), Proceed~\cite{zhao2025proceed}, $\delta$-Adapter~\cite{liang2026forecast}, TAFAS~\cite{kim2025tafas}, COSA~\cite{im2026cosa}, and SOLID~\cite{chen2024solid}, and cross-channel kNN-MTS~\cite{zhang2025knnmts} use the same iTransformer checkpoints without time features at $L = 96$, with only steps that overwrite issued forecasts disabled (Appendix~\ref{app:impl}).

\paragraph{Protocol.}
We report MSE (MAE in Appendix~\ref{app:results}) on every consecutive test origin, dropping no test batch.
Model selection, early stopping, and combination-weight estimation use only the validation segment; each forecaster is trained with three seeds and the results are averaged.
FreshCast uses one set of hyperparameters for all settings (sensitivity in Appendix~\ref{app:impl}).

\subsection{Generality}
\label{sec:generality}

Table~\ref{tab:generality} and Figure~\ref{fig:results}(a) report the results ($L = 336$ and 720: Table~\ref{tab:generality-long}). We have the following observations.

1) \textbf{FreshCast lowers the average error for every evaluated forecaster and input length.}
Across 812 settings, FreshCast reduces MSE in 782, falls back to the original forecaster with a zero weight in 20, and increases MSE in 10.
The mean reduction is 14.0\% over ten forecasters at $L = 96$ and 5.7\% over nine at $L = 720$, and for the original forecasters at each input length, both a paired test and a dataset-level test are significant (Appendix~\ref{app:results}).
Frozen foundation models also benefit: FreshCast lowers the MSE of Chronos-Bolt and Sundial by 9.7\% and 6.9\% (worse in 0 and 2 of 28 settings; Table~\ref{tab:generality-long}).

2) \textbf{The lowest-error forecaster on each dataset also benefits.}
At $L = 96$, the best forecaster on each dataset improves further, e.g., iTransformer on Traffic from 0.424 to 0.384 (Table~\ref{tab:generality}); in Figure~\ref{fig:intro}(a), the gain shrinks on stronger forecasters but stays positive.

3) \textbf{The gain shrinks with longer inputs but persists, and the role of the combination grows.}
The memory forecast alone reduces the error by 13.4\% at $L = 96$, close to the combination, but by only 3.6\% at $L = 720$, against 5.7\% for the combination (Appendix~\ref{app:results}).
Consistent with Observation~1 and Proposition~\ref{prop:chain}, this suggests that as periodic information becomes redundant, the memory shifts from substituting for the forecaster to complementing it, and the calibrated combination contributes a larger share.

\subsection{Comparison with Retrieval-Augmented and Online Methods}
\label{sec:comparison}

Table~\ref{tab:comparison} and Figure~\ref{fig:results}(b) report the comparisons.

1) \textbf{The gain of GTR largely vanishes with long inputs.}
FreshCast has lower MSE than GTR in 292 of 308 settings.
On MLP, DLinear, and iTransformer, GTR's error reduction falls from 3.8\% at $L = 96$ to 1.0\% and 0.7\% at 336 and 720, whereas FreshCast still lowers their error by 5.0\%--9.3\% there, consistent with Observation~1.

2) \textbf{Having the new observations is not enough.}
Relative to the original RAFT model, FreshCast has lower MSE in 28/28, 22/28, and 26/28 settings at the three input lengths, including RAFT's own $L = 720$ setting.
If the trained RAFT models are kept fixed and their bank grows under the same causal rule, RAFT improves by 3.1\% at $L = 96$ and not at 336 or 720, and FreshCast remains 10.1\%, 5.0\%, and 4.4\% lower (kNN-MTS with a growing bank: $+0.3\%$ on the frozen iTransformer, FreshCast 12.4\% lower): letting the memory grow is not sufficient by itself, and how new observations are read and adopted matters as well.

3) \textbf{The advantage also holds in the univariate setting.}
FreshCast has significantly lower MSE than PFRP on all four forecasters (seed-level paired tests, $p \le 3.7 \times 10^{-4}$) and is better on six of the seven datasets with SparseTSF, PFRP's main model.

4) \textbf{Writing deployment-time observations into parameters gains less than writing them into the memory.}
On the same forecaster, FreshCast has 7.2\%--13.8\% lower MSE than the six online updating methods, in at least 27 of 28 settings each ($p \le 2.9 \times 10^{-15}$), and daily fine-tuning of all parameters lowers the error by only 0.10\% (MLP and PatchTST; Appendix~\ref{app:ablation}).
All exceptions lie at ETTh2, $H = 720$, where FreshCast falls back to the forecaster (Appendix~\ref{app:limitations}).
They change the error by $-5.8\%$ to $+1.9\%$, against $-12.5\%$ for FreshCast.

\begin{table}[t]
\caption{Mean MSE change of FreshCast relative to each method in its comparison setting (negative: FreshCast lower; --: not evaluated). All baselines are rerun.}
\label{tab:comparison}
\footnotesize
\setlength{\tabcolsep}{0pt}
\begin{tabular*}{\columnwidth}{@{\extracolsep{\fill}}llccc}
\toprule
Method & Forecaster & $L = 96$ & $L = 336$ & $L = 720$ \\
\midrule
GTR & MLP & $-7.9\%$ & $-4.2\%$ & $-4.0\%$ \\
 & DLinear & $-15.1\%$ & $-7.6\%$ & $-8.6\%$ \\
 & PatchTST$^{\S}$ & $-26.9\%$ & $-21.9\%$ & -- \\
 & iTransformer & $-12.7\%$ & $-8.1\%$ & $-8.3\%$ \\
\midrule
RAFT (original) & MLP & $-12.8\%$ & $-5.1\%$ & $-4.1\%$ \\
RAFT, streaming$^{\ddagger}$ & MLP & $-10.1\%$ & $-5.0\%$ & $-4.4\%$ \\
\midrule
PFRP & SparseTSF & $-3.4\%$ & -- & -- \\
 & DLinear & $-7.0\%$ & -- & -- \\
 & PatchTST & $-4.8\%$ & -- & -- \\
 & TimesNet & $-4.6\%$ & -- & -- \\
\bottomrule
\end{tabular*}
\par\smallskip
\begin{tabular*}{\columnwidth}{@{\extracolsep{\fill}}lccccccc}
\toprule
 & DynaME & Proceed & $\delta$-Adapter & TAFAS & COSA & SOLID & kNN-MTS$^{\ddagger}$ \\
\midrule
$L = 96^{\dagger}$ & $-7.2\%$ & $-13.0\%$ & $-13.8\%$ & $-11.4\%$ & $-12.2\%$ & $-12.3\%$ & $-12.4\%$ \\
\bottomrule
\end{tabular*}
\par\smallskip\raggedright\scriptsize $^{\dagger}$The same iTransformer checkpoints without time features, as natively supported by DynaME. $^{\ddagger}$Bank grows at test time (Appendices~\ref{app:impl} and~\ref{app:ablation}). $^{\S}$Worse than PatchTST alone in 25/28 and 27/28 settings (Appendix~\ref{app:impl}); not in Figure~\ref{fig:intro}(a).
\end{table}

\subsection{Ablation}
\label{sec:ablation}

The upper block of Table~\ref{tab:ablation} crosses the two key determinants, timeliness (streaming or static memory) and complementarity (weight estimated on the validation or training segment); the lower block builds the memory forecast up from the simplest version.
The ablations use all channels of the seven datasets; MLP, iTransformer, CycleNet, and the five 2025--2026 forecasters; $L \in \{96, 720\}$; forecaster seed 2024; and kernel seeds 0--2.

1) \textbf{Most of the gain comes from the streaming memory.}
Freezing the memory at the end of the training segment removes almost all of the roughly 11\% error reduction of MLP, iTransformer, and CycleNet at $L = 96$ and most of that of the 2025--2026 forecasters, indicating that timeliness is a primary source of the gain.

2) \textbf{With long inputs, the weight is better estimated on validation data.}
With a streaming memory, a training-segment weight is smaller, about half of the validation estimate at $L = 720$ ($p = 4 \times 10^{-29}$), when the forecaster's training error is below its validation error (Appendix~\ref{app:results}).
The cell where both axes are unfavorable (static memory, training-segment weight) deviates most, about 13\% above the forecaster at $L = 96$, highlighting the interaction of the two factors (Theorem~\ref{thm:safety}(iii)).

3) \textbf{Shape analogs and nonlinear scoring work together.}
The untrained median of seasonal analogs already reduces the error at $L = 96$, but at $L = 720$ it is worse than the forecaster in 102 of 224 settings.
From linear scoring, switching to nonlinear scoring alone or adding shape analogs alone takes MLP at $L = 720$ only from $-3.2\%$ to $-3.5\%$ and $-3.6\%$, whereas both together reach $-4.9\%$; the joint increment is 1.5--2.4 times the sum of the single ones ($L = 720$, each forecaster).
This suggests that shape analogs from different times and levels need nonlinear scoring to be told apart.

4) \textbf{Timeliness matters more than weighting, yet learned weighting contributes materially.}
On the same streaming memory, the untrained rule selected per channel already beats FreshCast on a static memory, yet its MSE is about 5.7\% and 3.9\% higher than that of FreshCast at $L = 96$ and 720, and it is worse than the forecaster in 25 of the 168 settings of MLP, iTransformer, and CycleNet (FreshCast in none), all meeting the harm condition of Theorem~\ref{thm:safety}.

\begin{table}[t]
\caption{Ablation: MSE change (\%) relative to the forecaster; Recent: mean of the five 2025--2026 forecasters; superscripts: settings worse than the forecaster (of 28, or 140 for Recent). Rows 7--8 only switch to nonlinear scoring or only add shape analogs; the last row is the untrained rule.}
\label{tab:ablation}
\scriptsize
\setlength{\tabcolsep}{0pt}
\begin{tabular*}{\columnwidth}{@{\extracolsep{\fill}}l*{8}{c}}
\toprule
 & \multicolumn{4}{c}{$L = 96$} & \multicolumn{4}{c}{$L = 720$} \\
\cmidrule(lr){2-5}\cmidrule(lr){6-9}
Variant & MLP & iTr. & Cyc. & Recent & MLP & iTr. & Cyc. & Recent \\
\midrule
Streaming, val.\ $\what$ (FreshCast) & $-11.5$ & $-12.6$ & $-9.7$ & $-14.1$ & $-4.9$ & $-8.7$ & $-5.0$ & $-4.6^{6}$ \\
Streaming, train.\ $w_{\mathrm{train}}$ & $-11.2$ & $-12.3$ & $-9.3$ & $-14.4$ & $-3.8$ & $-5.6$ & $-3.5$ & $-3.9^{8}$ \\
Static, val.\ $\what$ & $+0.0^{8}$ & $-0.7^{6}$ & $+0.7^{8}$ & $-2.0^{46}$ & $+0.7^{8}$ & $-0.6^{6}$ & $+1.5^{9}$ & $+1.7^{53}$ \\
Static, train.\ $w_{\mathrm{train}}$ & $+12.7^{17}$ & $+13.7^{15}$ & $+13.4^{17}$ & $+15.4^{80}$ & $+2.4^{11}$ & $-2.0^{4}$ & $+1.8^{12}$ & $+8.6^{84}$ \\
\midrule
Seasonal, median (untrained) & $-4.7^{3}$ & $-5.5^{1}$ & $-4.3^{2}$ & $-6.9^{14}$ & $+0.1^{11}$ & $-1.7^{10}$ & $+0.4^{12}$ & $+1.3^{69}$ \\
Seasonal, linear scoring & $-10.3$ & $-11.4$ & $-9.2$ & $-12.9$ & $-3.2^{2}$ & $-6.4$ & $-3.4$ & $-2.4^{2}$ \\
Seasonal, nonlinear scoring & $-9.9$ & $-11.1$ & $-8.7$ & $-12.6$ & $-3.5^{2}$ & $-7.0$ & $-3.8$ & $-3.0^{6}$ \\
+ Shape analogs, linear & $-9.3$ & $-10.4$ & $-8.0$ & $-12.1$ & $-3.6^{2}$ & $-7.2$ & $-3.8$ & $-3.3^{2}$ \\
Both families, untrained rule & $-6.2^{1}$ & $-7.6^{1}$ & $-5.3^{1}$ & $-8.7^{7}$ & $-1.7^{8}$ & $-4.6^{5}$ & $-1.5^{9}$ & $-0.7^{51}$ \\
\bottomrule
\end{tabular*}
\end{table}

\subsection{Where Does the Improvement Come From?}
\label{sec:causal}

Under the evaluated controls, the improvement comes mainly from post-training observations in each channel's own memory.

\paragraph{Parameters and calibration, or retrieved content.}
With another channel's memory and unchanged kernel parameters, the gain disappears (weights below 0.08), and a calibrated combination with persistence changes the error within $\pm 0.4\%$ (weights below 0.13); the gain thus comes mainly from the channel's own history.

\paragraph{What remains beyond periodicity.}
Tiling the latest complete period in place of the memory, which mainly retains periodic structure, brings almost no improvement on Traffic and Electricity at $L = 720$, whereas the full memory still helps (Appendix~\ref{app:theory-checks}; Proposition~\ref{prop:chain}), and the gain does not grow with channel periodicity (Appendix~\ref{app:results}).
Excluding the last $H$ steps from the memory raises the error by under 2\%, so the gain does not rest on the latest observations alone.

\subsection{Failure Cases and Efficiency}
\label{sec:safety}
\label{sec:efficiency}

\paragraph{Failure cases.}
All 10 of the 812 settings worse than the forecaster involve the 2025--2026 forecasters and PatchTST with long inputs.
In the worst, CMoS on ETTh2 at $L = 720$ ($+3.0\%$), CMoS is worse than the memory on validation data ($\what = 0.87$) but better on test data (optimal weight 0.24), the case characterized by Theorem~\ref{thm:safety}(iii).
On the worst 0.1\% of windows, FreshCast has lower mean error than the forecaster in 31 of 36 settings and equal error in 3 (Appendix~\ref{app:theory-checks}).

\paragraph{Efficiency.}
On Traffic ($L = H = 96$), candidate generation, kernel training, and weight estimation take 3.65 minutes once for all forecasters, whereas GTR retrains the full model per forecaster (4.7 to 80.5 extra GPU minutes); inference per origin (862 channels) takes 1.25 s on one thread, far below the hourly sampling interval, with 1.16 GiB resident under bounded storage (RAFT's keys: 6.5 GiB).

\section{Conclusion}
\label{sec:conclusion}

We studied where retrieval gains come from for a forecaster frozen after deployment and identified two key determinants, timeliness and complementarity, whose relation can shift between validation and deployment as the memory becomes stale.
FreshCast streams the memory, weights references by relational kernel regression, and calibrates the correction by how much it explains the forecaster's residuals, reducing average error across ten forecasters, seven datasets, and three input lengths.
For a frozen forecaster, similarity alone is not enough: useful retrieval must stay timely and provide information that the forecaster still misses. Judging per query whether a reference can be trusted remains open (Appendix~\ref{app:limitations}).

\bibliographystyle{ACM-Reference-Format}
\bibliography{references}

\appendix
\section{Implementation and Protocol}
\label{app:impl}

\paragraph{Datasets.}
ETTh1 and ETTh2 (7 channels, hourly) and ETTm1 and ETTm2 (7 channels, 15 minutes) use 12/4/4-month splits (8{,}640/2{,}880/2{,}880 hourly steps, four times as many at 15 minutes); Weather (21 channels, 10 minutes), Electricity (321, hourly), and Traffic (862, hourly) use 7:1:2 splits of 52{,}696, 26{,}304, and 17{,}544 steps; daily and weekly periods follow from the interval. Each channel is standardized with its training-segment statistics, and multivariate metrics use FP16-rounded forecasts (univariate: FP32).

\paragraph{Forecasters and baselines.}
The five original forecasters use the official scripts of the public GTR code (long-input Traffic borrows Electricity's training hyperparameters); PatchTST uses its repository's official configuration at $L = 336$ and the same hyperparameters at $L = 96$, and the 2025--2026 forecasters their official configurations with at most 30 epochs and patience 5.
GTR uses its repository's scripts (GTR+PatchTST shares PatchTST's official configuration, as only an Electricity script exists), RAFT the settings of its Table~6, and PFRP rebuilds its memory bank and encoder from the training segment only, on our split instead of its 6:2:2 ETT split.
Chronos-Bolt and Sundial (base sizes) forecast each channel once for 720 steps per origin, truncated to each $H$, taking medians (Sundial: 20 samples); pretraining may have covered Electricity (Chronos-Bolt) and Traffic (Sundial), which affects only the forecasters' errors.

\paragraph{Online updating methods.}
The seven methods and FreshCast use the same checkpoints (an iTransformer without time features, otherwise as in the main table; three seeds), and each origin uses only observations whose full future has arrived.
DynaME uses the public code (commit 0743ad6) with hyperparameters selected on validation data; the original DynaME, which also updates the forecaster, is not significantly better than the frozen variant (17 of 28 settings; $-0.5\%$, $p = 0.13$).
Proceed, $\delta$-Adapter, TAFAS, COSA, and SOLID use the recommended configurations of their official scripts (Proceed: OnlineTSF's iTransformer settings; $\delta$-Adapter: multiplicative form).
kNN-MTS retrieves across channels as in its code, with temperature and mixing scale set on validation data.

\paragraph{Settings of FreshCast.}
Per channel, the memory stores the standardized raw series and each window's mean, standard deviation, and normalized shape vector; $L_q = 96$.
Seasonal analogs go back $J = 8$ daily and weekly periods within the most recent $K = 112$ days (at most 16), each using the next $\min(H, P)$ observed steps, tiled when $H$ exceeds the period; a static memory takes them by phase from its last $K$ days and keeps the same scoring function.
The shape-analog pool holds all admissible windows of the recent $K$ days and $Q = 500$ earlier windows with fixed random indices, from which the $N = 200$ most similar are taken (about 216 candidates per query); window standard deviations are floored at 0.01 and continuations clipped to $\pm 1000$ so that Eq.~\eqref{eq:reanchor} does not amplify near-constant windows.
The 11 descriptors are two family flags, shape similarity, age, two phase distances, the level gap $|\hat{\mu}_c - \hat{\mu}_q| / \hat{\sigma}_q$, the scale ratio $\log(\hat{\sigma}_c / \hat{\sigma}_q)$, and three training-segment statistics (daily and weekly autocorrelation, mean squared difference); the scoring function has 5{,}006 parameters, the linear-scoring rows of Table~\ref{tab:ablation} use the same descriptors, and the nonlinear seasonal row retrains FreshCast's scoring function on seasonal candidates.
Relational kernel regression is trained on 209 channels (all of ETT and Weather and 80 seeded channels each of Electricity and Traffic) with 150 random training-segment origins per channel, $c_{\max} = 20$, AdamW (learning rate $2 \times 10^{-3}$, weight decay $10^{-2}$), batch size 256, and 30 epochs with cosine decay and a parameter moving average (0.995); the main tables use five seeds.
The untrained rule mixes, with fixed weights, the median of same-phase values over the last $M$ periods and the mean of the $k$ most similar shape analogs, with the period, $M$, $k$, and the weight selected per channel on the validation segment.

\paragraph{Sensitivity.}
All settings are shared across datasets, forecasters, and input lengths, and the most influential one, the combination weight, is closed-form. Table~\ref{tab:sensitivity} changes one setting at a time as in the ablations, retraining the kernel and rebuilding the memory.

\begin{table}[h]
\caption{Difference in MSE change vs.\ the forecaster (points; positive: worse) when one setting is changed; mean over three forecasters, seven datasets, four horizons.}
\label{tab:sensitivity}
\scriptsize
\setlength{\tabcolsep}{2pt}
\renewcommand{\arraystretch}{0.85}
\begin{tabular*}{\columnwidth}{@{\extracolsep{\fill}}lccc}
\toprule
Setting (FreshCast) & Changed to & $L = 96$ & $L = 720$ \\
\midrule
Query length $L_q$ (96) & 336 / 720 & $+0.64$ / $+1.42$ & $+0.97$ / $+1.42$ \\
Recent window $K$ (112 days) & 56 / 28 & $-0.09$ / $+0.07$ & $-0.05$ / $-0.11$ \\
Periods back $J$ (8) & 4 / 16 & $+0.01$ / $-0.30$ & $-0.33$ / $-0.30$ \\
Shape analogs $N$ (200) & 25 / 50 / 100 & $-0.02$ / $+0.05$ / $-0.10$ & $-0.04$ / $-0.01$ / $-0.07$ \\
Distant windows $Q$ (500) & 0 & $-0.00$ & $+0.20$ \\
Hidden width (64) & 32 / 128 & $+0.25$ / $-0.10$ & $+0.18$ / $-0.03$ \\
$\varepsilon_c$ segment (validation) & training & $-0.18$ & $-0.17$ \\
\bottomrule
\end{tabular*}
\end{table}

Except for longer queries, which do not restore the long-input gain, mean changes stay within 0.35 points with dataset-dependent signs; distant windows matter only for long inputs, nonlinear scoring does not gain from width, and a training-segment $\varepsilon_c$ stays within seed noise (the main results keep the validation estimate).

\paragraph{Bounded storage, transfer, and hardware.}
As distant windows have fixed random indices, storing the recent $K$ days and the $Q$ distant windows with the smallest indices suffices; on 10 channels of the seven datasets (about 250{,}000 queries), the test memory forecast differs in relative error by at most $3 \times 10^{-5}$, and on Traffic the resident memory is 1.16 GiB at $K = 112$ days and 0.72 GiB at 56 days, of which only the raw series grows.
A kernel trained on six datasets and applied to the seventh gives a mean reduction of $-8.4\%$, against $-9.1\%$ with all seven (eight forecasters, $L \in \{96, 720\}$; 8 of 448 settings worse than the forecaster), and Traffic changes most ($-13.0\%$ to $-11.1\%$), as holding it out removes nearly 40\% of the training channels.
We use NVIDIA L20 GPUs, PyTorch 2.4.1, and CUDA 12.1; with 48 CPU processes, Traffic takes 0.051 s per origin.

\section{Proofs}
\label{app:proofs}

\paragraph{Proposition~\ref{prop:chain}.}
The decomposition is the chain rule of mutual information, and conditional independence gives $I(Y; M_p \mid X_L) = 0$.

\paragraph{Proposition~\ref{prop:staleness}.}
(i) For $\tau$ a multiple of $P$, $y_{t+k} - y^{+}_{t+k} = \Delta m + \Delta a\, \psi(\varphi_{t+k}) + \varepsilon_{t+k} - \varepsilon_{t+k-\tau}$ with $\Delta m = m_{t+k} - m_{t+k-\tau}$ and $\Delta a = a_{t+k} - a_{t+k-\tau}$; the cross terms vanish by independence, $\mathbb{E}[\psi] = 0$, and $\mathbb{E}[\psi^2] = 1$, giving $D(\tau) + A(\tau) + 2\sigma_\varepsilon^2$.
(ii) With $(m_c, a_c)$ and $(m_q, a_q)$ constant in the candidate and query windows and window variances $s^2 = a^2 + \sigma_\varepsilon^2$, substituting into Eq.~\eqref{eq:reanchor} gives, to first order in the estimation errors of the window statistics, $\hat{y} - y = (a_c s_q / s_c - a_q)\,\psi + (s_q / s_c)\,\varepsilon^{c} - \varepsilon^{q}$ plus these errors: $m_q - m_c$ cancels, and the coefficient of $\psi$ vanishes when $a_c = a_q$ or as $\sigma_\varepsilon^2 / a^2 \to 0$.

\paragraph{Corollary~\ref{cor:static}.}
In a static memory, the continuation of the newest segment ends at $T_{\mathrm{tr}}$ and its past window at $T_{\mathrm{tr}} - H$, so the minimum age at origin $t$ is $t - T_{\mathrm{tr}} + H$; a streaming memory writes in every window satisfying Definition~\ref{def:causal}, so its minimum age is $H$.
For a stationary, mixing Markov level process, $m_{t-\tau_2} \to m_{t-\tau_1} \to m_t$ ($\tau_1 \le \tau_2$) is a Markov chain, so the data processing inequality makes $I(m_t; m_{t-\tau})$ non-increasing in $\tau$ and mixing drives it to zero.

\paragraph{Proposition~\ref{prop:gain}.}
Completing the square gives $\mathrm{MSE}(w) = S_{ee} - 2wS_{ed} + w^2 S_{dd} = S_{ee}(1 - \rho^2) + S_{dd}(w - w^{*})^2$.
If $0 \le w^{*} \le 1$, the minimum is at $w^{*}$ with reduction $\rho^2 = \kappa^2/\delta$; if $S_{ed} \le 0$, $\mathrm{MSE}$ is non-decreasing on $[0, 1]$.

\paragraph{Theorem~\ref{thm:safety}.}
(i) By Proposition~\ref{prop:gain}, $\mathrm{MSE}(\what) - S_{ee} = S_{dd}(\what - w^{*})^2 - S_{ee}\rho^2$, and $S_{ee}\rho^2 = S_{dd} w^{*2}$.
(ii) With $v, b$ and $v_q, b_q$ the variances and bounds of $z = d(e - w^{*} d)$ and $d^2 - S_{dd}$,
\begin{align*}
  \Pr\bigl(\mathrm{MSE}(\what) > S_{ee}\bigr) \le{} & 2\exp\!\left( -\frac{n\, \kappa^2 S_{ee}^2}{8v + \tfrac{4}{3}\, b\, \kappa\, S_{ee}} \right) \\
  & + \exp\!\left( -\frac{n\, S_{dd}^2}{8 v_q + \tfrac{4}{3}\, b_q S_{dd}} \right).
\end{align*}
The unclipped estimate $\tilde{w} = \hat{S}_{ed}/\hat{S}_{dd}$ satisfies $\tilde{w} - w^{*} = \bar{z}/\hat{S}_{dd}$ with $\mathbb{E}[z] = 0$, and for $w^{*} \in (0, 1]$ clipping only decreases $|\what - w^{*}|$. On the event $\hat{S}_{dd} \ge S_{dd}/2$, $|\what - w^{*}| > w^{*}$ implies $|\bar{z}| > S_{ed}/2$, so the harm probability is at most $\Pr(|\bar{z}| \ge \tfrac{1}{2}\kappa S_{ee}) + \Pr(\hat{S}_{dd} - S_{dd} < -\tfrac{1}{2} S_{dd})$, and the two terms follow from the two-sided and one-sided Bernstein inequalities.
(iii) $\what$ depends only on the validation segment; repeating (i) under the test distribution gives the result.

\section{Theoretical Checks}
\label{app:theory-checks}

Units are 129 diagnostic channels $\times$ 4 horizons $\times$ each forecaster $\times$ 3 kernel seeds, with one forecaster seed.

\begin{table*}[t]
\caption{Test MSE of the ten forecasters alone (Base) and with FreshCast at $L = 336$ and 720 (mean over four horizons; PatchTST not run at 720). Bottom: two foundation models, zero-shot with a 512-step context.}
\label{tab:generality-long}
\scriptsize
\setlength{\tabcolsep}{0pt}
\renewcommand{\arraystretch}{0.8}
\begin{tabular*}{\textwidth}{@{\extracolsep{\fill}}ll*{10}{c}}
\toprule
 & & \multicolumn{2}{c}{MLP} & \multicolumn{2}{c}{DLinear} & \multicolumn{2}{c}{PatchTST} & \multicolumn{2}{c}{iTransformer} & \multicolumn{2}{c}{CycleNet} \\
\cmidrule(lr){3-4}\cmidrule(lr){5-6}\cmidrule(lr){7-8}\cmidrule(lr){9-10}\cmidrule(lr){11-12}
Dataset & $L$ & Base & +FreshCast & Base & +FreshCast & Base & +FreshCast & Base & +FreshCast & Base & +FreshCast \\
\midrule
ETTh1 & 336 & 0.430 & 0.404 & 0.427 & 0.424 & 0.425 & 0.408 & 0.463 & 0.412 & 0.439 & 0.412 \\
 & 720 & 0.441 & 0.403 & 0.442 & 0.426 & -- & -- & 0.485 & 0.410 & 0.440 & 0.401 \\
ETTh2 & 336 & 0.373 & 0.361 & 0.476 & 0.369 & 0.344 & 0.344 & 0.376 & 0.360 & 0.374 & 0.362 \\
 & 720 & 0.362 & 0.357 & 0.552 & 0.363 & -- & -- & 0.386 & 0.365 & 0.367 & 0.359 \\
ETTm1 & 336 & 0.362 & 0.330 & 0.358 & 0.339 & 0.352 & 0.328 & 0.370 & 0.333 & 0.362 & 0.329 \\
 & 720 & 0.360 & 0.334 & 0.357 & 0.341 & -- & -- & 0.375 & 0.334 & 0.362 & 0.332 \\
ETTm2 & 336 & 0.267 & 0.249 & 0.287 & 0.247 & 0.257 & 0.246 & 0.279 & 0.249 & 0.270 & 0.250 \\
 & 720 & 0.266 & 0.248 & 0.269 & 0.245 & -- & -- & 0.288 & 0.250 & 0.263 & 0.248 \\
Weather & 336 & 0.234 & 0.218 & 0.245 & 0.223 & 0.230 & 0.217 & 0.239 & 0.220 & 0.226 & 0.215 \\
 & 720 & 0.234 & 0.221 & 0.237 & 0.224 & -- & -- & 0.237 & 0.222 & 0.224 & 0.217 \\
Electricity & 336 & 0.162 & 0.155 & 0.167 & 0.156 & 0.163 & 0.156 & 0.163 & 0.151 & 0.158 & 0.153 \\
 & 720 & 0.160 & 0.154 & 0.160 & 0.155 & -- & -- & 0.163 & 0.151 & 0.157 & 0.153 \\
Traffic & 336 & 0.397 & 0.384 & 0.450 & 0.398 & 0.410 & 0.390 & 0.392 & 0.378 & 0.413 & 0.387 \\
 & 720 & 0.391 & 0.384 & 0.443 & 0.396 & -- & -- & 0.387 & 0.376 & 0.400 & 0.386 \\
\midrule
 & & \multicolumn{2}{c}{TQNet} & \multicolumn{2}{c}{XLinear} & \multicolumn{2}{c}{MixLinear} & \multicolumn{2}{c}{TimeBase} & \multicolumn{2}{c}{CMoS} \\
\cmidrule(lr){3-4}\cmidrule(lr){5-6}\cmidrule(lr){7-8}\cmidrule(lr){9-10}\cmidrule(lr){11-12}
Dataset & $L$ & Base & +FreshCast & Base & +FreshCast & Base & +FreshCast & Base & +FreshCast & Base & +FreshCast \\
\midrule
ETTh1 & 336 & 0.427 & 0.407 & 0.427 & 0.408 & 0.414 & 0.398 & 0.409 & 0.399 & 0.416 & 0.399 \\
 & 720 & 0.442 & 0.401 & 0.431 & 0.402 & 0.404 & 0.392 & 0.411 & 0.402 & 0.419 & 0.399 \\
ETTh2 & 336 & 0.362 & 0.353 & 0.351 & 0.350 & 0.349 & 0.349 & 0.358 & 0.355 & 0.352 & 0.352 \\
 & 720 & 0.365 & 0.353 & 0.355 & 0.354 & 0.341 & 0.343 & 0.358 & 0.355 & 0.350 & 0.352 \\
ETTm1 & 336 & 0.357 & 0.330 & 0.346 & 0.327 & 0.377 & 0.333 & 0.368 & 0.331 & 0.353 & 0.328 \\
 & 720 & 0.361 & 0.332 & 0.348 & 0.330 & 0.379 & 0.335 & 0.363 & 0.334 & 0.360 & 0.332 \\
ETTm2 & 336 & 0.262 & 0.246 & 0.258 & 0.246 & 0.260 & 0.248 & 0.259 & 0.248 & 0.260 & 0.249 \\
 & 720 & 0.264 & 0.247 & 0.258 & 0.247 & 0.257 & 0.247 & 0.252 & 0.246 & 0.258 & 0.249 \\
Weather & 336 & 0.232 & 0.217 & 0.225 & 0.215 & 0.250 & 0.224 & 0.225 & 0.216 & 0.227 & 0.217 \\
 & 720 & 0.238 & 0.223 & 0.228 & 0.220 & 0.241 & 0.224 & 0.219 & 0.215 & 0.223 & 0.217 \\
Electricity & 336 & 0.165 & 0.153 & 0.166 & 0.156 & 0.178 & 0.159 & 0.177 & 0.159 & 0.167 & 0.158 \\
 & 720 & 0.167 & 0.155 & 0.157 & 0.150 & 0.171 & 0.159 & 0.169 & 0.158 & 0.161 & 0.156 \\
Traffic & 336 & 0.411 & 0.385 & 0.416 & 0.390 & 0.440 & 0.399 & 0.447 & 0.399 & 0.424 & 0.395 \\
 & 720 & 0.418 & 0.394 & 0.406 & 0.389 & 0.417 & 0.396 & 0.419 & 0.396 & 0.412 & 0.394 \\
\bottomrule
\end{tabular*}
\par\smallskip
\begin{tabular*}{\textwidth}{@{\extracolsep{\fill}}l*{8}{c}}
\toprule
Base / +FreshCast & ETTh1 & ETTh2 & ETTm1 & ETTm2 & Weather & Electricity & Traffic & Mean change \\
\midrule
Chronos-Bolt & 0.448 / 0.423 & 0.367 / 0.361 & 0.421 / 0.344 & 0.291 / 0.256 & 0.263 / 0.218 & 0.159 / 0.152 & 0.435 / 0.387 & $-9.7\%$ \\
Sundial & 0.402 / 0.392 & 0.344 / 0.345 & 0.371 / 0.331 & 0.277 / 0.252 & 0.243 / 0.219 & 0.148 / 0.144 & 0.460 / 0.393 & $-6.9\%$ \\
\bottomrule
\end{tabular*}
\end{table*}

\begin{table*}[t]
\caption{Univariate test MSE of the PFRP paper's forecasters alone (Base), with PFRP, and with FreshCast ($L = 96$; its target channels; mean over horizons; best in bold).}
\label{tab:pfrp}
\scriptsize
\setlength{\tabcolsep}{0pt}
\renewcommand{\arraystretch}{0.8}
\begin{tabular*}{\textwidth}{@{\extracolsep{\fill}}l*{12}{c}}
\toprule
 & \multicolumn{3}{c}{SparseTSF} & \multicolumn{3}{c}{DLinear} & \multicolumn{3}{c}{PatchTST} & \multicolumn{3}{c}{TimesNet} \\
\cmidrule(lr){2-4}\cmidrule(lr){5-7}\cmidrule(lr){8-10}\cmidrule(lr){11-13}
Dataset & Base & +PFRP & +FreshCast & Base & +PFRP & +FreshCast & Base & +PFRP & +FreshCast & Base & +PFRP & +FreshCast \\
\midrule
ETTh1 & 0.084 & 0.078 & \textbf{0.075} & 0.116 & 0.112 & \textbf{0.106} & 0.079 & 0.079 & \textbf{0.076} & 0.077 & 0.079 & \textbf{0.075} \\
ETTh2 & 0.202 & \textbf{0.193} & \textbf{0.193} & 0.224 & 0.214 & \textbf{0.208} & 0.201 & 0.191 & \textbf{0.187} & 0.191 & 0.192 & \textbf{0.186} \\
ETTm1 & 0.054 & 0.052 & \textbf{0.050} & 0.065 & 0.064 & \textbf{0.051} & 0.054 & 0.053 & \textbf{0.050} & 0.053 & 0.053 & \textbf{0.050} \\
ETTm2 & 0.126 & 0.120 & \textbf{0.117} & 0.126 & 0.122 & \textbf{0.116} & 0.123 & 0.120 & \textbf{0.114} & 0.122 & 0.120 & \textbf{0.115} \\
Weather & 0.0015 & 0.0015 & \textbf{0.0014} & 0.0062 & 0.0051 & \textbf{0.0042} & 0.0019 & \textbf{0.0017} & 0.0018 & 0.0017 & 0.0017 & \textbf{0.0016} \\
Electricity & 0.473 & 0.363 & \textbf{0.356} & 0.393 & 0.372 & \textbf{0.347} & 0.408 & 0.404 & \textbf{0.328} & 0.417 & 0.402 & \textbf{0.337} \\
Traffic & 0.239 & 0.209 & \textbf{0.191} & 0.278 & \textbf{0.177} & 0.192 & 0.178 & 0.175 & \textbf{0.168} & 0.220 & \textbf{0.179} & 0.191 \\
\bottomrule
\end{tabular*}
\end{table*}

\paragraph{Periodic tiling stops helping at long inputs on Traffic and Electricity (Proposition~\ref{prop:chain}).}
With a periodic tiling in place of the memory, the reduction on Traffic and Electricity falls from 2.5\%--7.9\% at $L = 96$ to $-0.1\%$ and $0.8\%$ at $L = 720$ (weights 0.03--0.13), while the full memory still gains 1.8\%--7.1\%; on ETT, $L = 96$ already covers the daily period and the tiling carries the latest values, so it still helps.

\paragraph{The error growth of value readouts comes mainly from the level term (Proposition~\ref{prop:staleness}).}
On the univariate targets of six datasets, 81\%--99\% of the age-related error growth of value-read seasonal analogs comes from the level term.
Over about 8.3 million paired readouts, from under two days to 16--32 weeks of age, the median value-readout error rises from 0.37 to 1.22 (per-channel units) while the shape readout stays within 0.40--0.48, and the shape win rate rises from 27\% to 72\%, crossing at 2--4 weeks; on Traffic (level gap below 0.25) the value readout wins at every age.
As 1\%--3\% of shape readouts are amplified over tenfold, this holds for medians and win rates but not means.
With a streaming memory, reading seasonal analogs by value only beats offering both readouts by 0.1\%--0.2\% (with a static memory it is 0.4\%--0.6\% worse at $L = 96$).

\paragraph{The validation-estimated combination gain correlates with the realized gain (Proposition~\ref{prop:gain}).}
The validation-estimated optimal gain has Spearman correlation 0.72, 0.52, and 0.47 with the test reduction at the three input lengths. Per channel, horizon, and seed, lower-error forecasters have lower complementarity (mean Spearman 0.84) and smaller reductions (0.92).
Over all units, complementarity correlates best with the test reduction among validation indicators (0.73, 0.56, 0.50), ahead of the memory-to-forecaster error ratio ($-0.65$, $-0.38$, $-0.33$), retrieval similarity (0.12--0.23), periodicity ($-0.09$ to $+0.01$), and either model's error alone ($|\cdot| \le 0.13$).

\paragraph{Every harm corresponds to a shift of the weight (Theorem~\ref{thm:safety}).}
Under i.i.d.\ resampling, the log harm probability correlates with $n\kappa^2$ at $-0.57$, $-0.58$, and $-0.65$. The validation $\what$ harms 3.9\%, 5.5\%, and 6.6\% of test units, all with $|\what - w^{*}_{\mathrm{te}}| > w^{*}_{\mathrm{te}}$; at $L = 96$ the harm rate falls from 16.9\% for $\kappa$ in 0--0.05 to 1.7\% above 0.3.

\paragraph{Two distribution shifts.}
With a static memory, the validation $\what$ exceeds the test optimum in 17--26 of 28 settings and the optimum of 48 of 56 equal test intervals, and static FreshCast is worse than the forecaster in some intervals on six datasets, whereas streaming FreshCast improves in every interval of every dataset.
The untrained rule matches relational kernel regression in validation complementarity (0.136 vs.\ 0.134 at $L = 96$) but has lower test complementarity in 84 of 84 ($L = 96$) and 82 of 84 ($L = 720$) settings.
Per window ($L = 96$; three forecasters, six channels per dataset), the worst 0.1\% of windows are worse only for CycleNet on Traffic and Electricity (6\% and 1\%).

\section{Obtaining New Observations Is Not Enough for the Same Gain}
\label{app:ablation}

A natural concern is that any method given the deployment-time observations would gain as much.
Online fine-tuning updates all forecaster parameters daily on the last 14 days of ground truth, with one of six configurations chosen on the validation segment and no update if it does not help there (74 of 168 pairs; MLP and PatchTST, seven datasets, four horizons, three seeds).
For RAFT with a streaming bank, the RAFT models of Table~\ref{tab:comparison} (its Table~6 settings, seeds 0--2) are kept fixed with their best-validation checkpoints, and at test time the bank holds every window whose full future has been observed, as in FreshCast; restricting it to training windows reproduces the static results exactly.

\section{Complete Results}
\label{app:results}

Table~\ref{tab:pfrp} compares with PFRP per dataset.
In MAE, FreshCast lowers the average error of every forecaster, dataset, and input length, beats GTR in 301 of 308 settings, RAFT in 80 of 84, and PFRP on three of four forecasters (a tie on SparseTSF), and each method in the lower block of Table~\ref{tab:comparison} in at least 27 of 28 settings.
For the original forecasters, paired tests over (dataset, $H$, forecaster, seed) units give $p \le 10^{-50}$ at each input length, and all seven datasets improve at every length (exact Wilcoxon signed-rank test, $p = 0.016$, the minimum for seven units).
The memory forecast alone ($w = 1$) reduces the error by 13.4\%, 3.9\%, and 3.6\% at $L = 96$, 336, and 720 and is worse than the forecaster in 6/280, 56/280, and 66/252 settings; the combination beats both in 191/280, 226/280, and 212/252, and the mean $\what$ (0.62, 0.49, 0.49) stays below the test optimum (0.78, 0.61, 0.60). Only MixLinear, TimeBase, and CMoS (training error 1.14--1.16 times validation error) gain from $w_{\mathrm{train}}$, by 0.7\%--0.8\% at $L = 96$.
By quintiles of periodicity (autocorrelation), the diagnostic channels improve by 12.5\%--20.7\% at $L = 96$ with no trend (Spearman $-0.09$ to $+0.01$), and the least periodic Weather by 16.1\%.
\section{Shape Analogs Help More on Irregular Events}
\label{app:synthetic}

As our target applications are Web services, we use 12 ten-minute request-load series of a regional Azure OpenAI service from BurstGPT~\cite{wang2025burstgpt} (API and conversation calls, with weak and strong periodicity) and insert real bursts from another period of the same service, scaled by the training standard deviation, controlling only their timing: one burst in the training segment and a different one afterwards, recurring at random times.
Against a seasonal-only variant (eight forecasters, $L = 96$ and 720, kernel trained on these series), shape analogs lower the event-window error by 3.8 and 4.2 points on API series and 2.2 and 2.7 on conversation series for all eight forecasters, but only by 0.6 and 1.2 when each event is a different burst, and by 2.2 and 3.4 when events occur at a fixed weekly time; FreshCast reduces the error under every condition.

\section{Limitations}
\label{app:limitations}

(1) The weight is set per dataset; an updated memory cannot remove changes in the relative reliability of forecaster and memory (Section~\ref{sec:safety}), so FreshCast gives no per-sample reliability guarantee.
(2) Where the level oscillates seasonally (ETTh2), the most useful segments lie a year earlier, beyond the recent window; keeping distant segments by season is a direct remedy.
(3) Memory grows linearly with the number of channels (1.16 GiB on Traffic).
(4) The theory assumes an additive, per-channel model and, in Theorem~\ref{thm:safety}(ii), i.i.d.\ samples, whereas windows are dependent.

\end{document}